# A Pareto Optimal D* Search Algorithm for Multiobjective Path Planning

Alexander Lavin

*Abstract*—Path planning is one of the most vital elements of mobile robotics, providing the agent with a collision-free route through the workspace. The global path plan can be calculated with a variety of informed search algorithms, most notably the A* search method, guaranteed to deliver a complete and optimal solution that minimizes the path cost. D* is widely used for its dynamic replanning capabilities. Path planning optimization typically looks to minimize the distance traversed from start to goal, but many mobile robot applications call for additional path planning objectives, presenting a multiobjective optimization (MOO) problem. Common search algorithms, e.g. A* and D*, are not well suited for MOO problems, yielding suboptimal results. The search algorithm presented in this paper is designed for optimal MOO path planning. The algorithm incorporates Pareto optimality into D*, and is thus named D*-PO. Non-dominated solution paths are guaranteed by calculating the Pareto front at each search step. Simulations were run to model a planetary exploration rover in a Mars environment, with five path costs. The results show the new, Pareto optimal D*-PO outperforms the traditional A* and D* algorithms for MOO path planning.

## I. INTRODUCTION

A crucial task for mobile robots is to navigate intelligently through their environment. It can be argued that path planning is one of the most important issues in the navigation process [1], and subsequently much research in field robotics is concerned with path planning [2], [3]. To complete the navigation task, methods will read the map of the environment and search algorithms will attempt to find free paths for the robot to traverse. Path planning methods find a path connecting the defined start and goal positions, while environmental parameters play the role as algorithm inputs, and the output is an optimized path from the start to goal [4]. The important issue in mobile robot navigation is optimizing path efficiency according to some parameters such as cost, distance, energy, and time. A* search is the leading search algorithm for mobile robot path planning, yielding solutions guaranteed to be optimal [5]. Recently the mobile robot community has put an increased emphasis on suboptimal path planning methods which meet time-critical constraints over slow, optimal algorithms. D* is the dynamic version of A*, sacrificing optimality for computational efficiency in changing environments. Notable implementations of D* in the field robotics include the DARPA Urban Challenge vehicle from Carnegie Mellon University, and Mars rovers Opportunity and Spirit [6].

Path planning methods typically optimize the path efficiency for one criterion, yet many mobile robot operations call for a path plan that optimizes for several parameters [7]. Path optimization over several parameters – e.g. distance and energy – is a multiobjective optimization (MOO) problem. The best path is not necessarily the shortest path, nor the path calling for the least amount of energy expenditure.

Combining the optimization criteria into a single objective function is a common approach, often with tools such as thresholds and penalty functions, and weights for linear combinations of objective values. But these methods are problematic as the final solution is typically very sensitive to small adjustments in the penalty function coefficients and weighting factors [8], posing an issue for MOO path planning with A*, D*, or similar search algorithms. Another issue is these methods yield suboptimal solution paths. Evolutionary algorithms, particularly genetic algorithms, have been used widely for MOO problems, including success in path planning [7], [9]. Drawing on evolutionary computation techniques in a previous study, Lavin [10] defined an A* alternative specifically designed for MOO path planning. This algorithm, named *A\*-PO*, yields *Pareto optimal* solutions: solutions in which there exist no other solutions superior in all objectives. The key feature in A*-PO is the use of a Pareto front cost function when searching for the solution path. Lavin shows A*-PO outperforms the standard A* search algorithm across all optimization criteria of a given MOO problem, with minimal loss to computational efficiency.

This paper continues Lavin's previous work of implementing Pareto optimality into multiobjective optimization for mobile robot path planning. Presented in this study is *D\*-PO*: a Pareto optimal version of the D* search algorithm specifically designed for MOO path planning. The path planning algorithms D*-PO and A*-PO are novel because each step is Pareto optimal, solving the aforementioned issue of suboptimal results.

The next section further discusses Pareto optimality and the application to mobile robot path planning. Section III discusses the technical approach used in this study, and Section IV presents the results. The results are over a set of simulations running the A*, D*, and D*-PO algorithms in Mars environments. Section V concludes the paper with discussion and future work.

## II. BACKGROUND AND METHODS

### A. Mobile Robot Path Planning

The aim of mobile robot path planning is to provide an efficient path from start to goal that avoids objects and obstacles. An efficient path is one that minimizes path costs, where the cost is typically the travel distance or time. Search algorithms find a negotiable path from start to goal states in the *configuration space*: data structures that show the position and orientations of objects and robots in the workspace area, including both the free and obstructed regions. Typically the robot's world is represented by either an occupancy grid, a vertex graph, or a Voronoi diagram [1]. The methods discussed in this study use an *occupancy grid* representation composed of binary states, where 0s and 1s represent free and occupied cells, respectively. As shown in the first flowchart block of Fig. 1, the robot (agent) has at most eight possible successors for the next step in the path; the robot in the green position is capable of moving into a neighboring yellow position, but not the occupied grey cells.

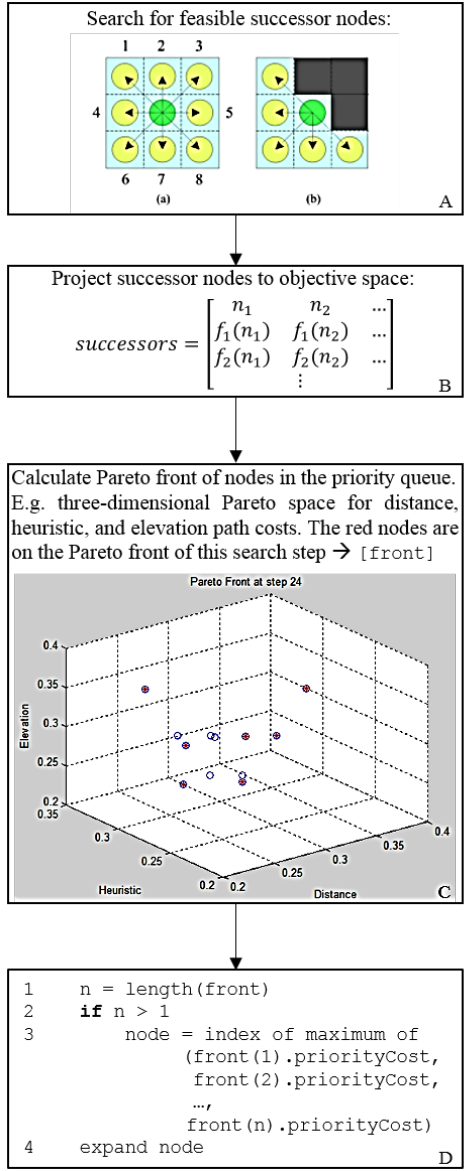

Fig. 1. High-level flow of using Pareto optimal path cost. This process is repeated at each step of the search algorithm. The algorithm first identifies feasible next nodes from its current location (1A). In 1B, the resulting matrix *successors* is the priority queue, which is normalized along the rows; more detail in Section III-A. In 1C, notice multiple nodes are on the Pareto front. The pseudocode of 1D decides which node to choose for expansion – i.e. the node with the higher priorityCost. This tiebreaker methodology is discussed more Section II-B.

Potential solution paths connect the start cell to the goal cell via free cells. Searching for the optimum path is an optimization problem, where the optimum path is defined as that which minimizes the path cost, or the *objective function*.

A candidate path can be denoted by

$$P = \{p_1, p_2, \ldots, p_n\} \quad (1)$$

where $p_i$ is the $i$th waypoint of the path $P$. The MOO problem is then framed as determining a path

$$P^* \in P \quad (2)$$

that satisfies

$$F(P^*) = min\{F_1(P), F_2(P), \ldots, F_t(P)\} \quad (3)$$

where $F_i$ denotes the $i$th cost function of the path planning problem. That is, equations (1-3) state the aim of the problem is to minimize path cost across all $t$ objective functions. The study here considers four cost functions, or $t = 4$. Equation (4) gives the total length of the path:

$$F_1(P) = \sum_{i=1}^{n-1} |p_i, p_{i+1}| \quad (4)$$

where $|p_i, p_{i+1}|$ is the Euclidean distance between subsequent cells in the path. Minimizing $F_1$ finds the path of shortest length from start to goal.

Equation (5) gives the average elevation of the path:

$$F_2(P) = \sum_{i=1}^{n} e_i/n \quad (5)$$

where $e_i$ is the elevation at waypoint $i \ldots n$. Running algorithms on workspaces with the same terrains and start/goal states, the minimization of $F_2$ gives the path which climbs up the least amount of incline (or alternatively moves the robot down the most decline).

The simulations in this study model a planetary exploration rover, thus a relevant path cost is solar vector deviation. That is, the MOO problem includes an additional optimization objective to minimize the total angular deviation of sunlight from the rover's rear solar panel. This was computed by minimizing the dot product of the rover vector $\vec{r}$ and the solar ray vector $\vec{s}$:

$$F_3(P) = \sum_{i=1}^{n} \vec{r_i} \cdot \vec{s_i} = |r_i||s_i|cos\theta \quad (6)$$

This cost layer is dependent on both time and the robot's orientation in the configuration space, and is thus it is dynamic. That is, the costmap changes at each step in the path, depending on the two-dimensional rover vector and the solar vector field. For this case study, the solar incidence was modeled as a two-dimensional vector whose angle rotated 0.01 radians per search step.

In the simulations an area of high risk in each workspace is to be avoided. Equation (7) models a cost for path risk:

$$F_4(P) = \frac{1}{n} \sum_{i=1}^{n} \frac{1}{(x-x_R)^2+(y-y_R)^2} \qquad (7)$$

The path risk decreases quadratically with the distance from the location of risk in the workspace. The costs of (4)-(7) are illustrated in Fig. 4. The aim is to minimize these four cost sums.

In path optimization the search method aims to plan an efficient path according to (3). With *a priori* knowledge of the exploration area, we can use an *informed search* method. Specifically *best-first search*, which traverses a graph or grid using a priority queue to find the shortest, collision-free path [4]. A *priority queue*, or *open list*, is used to focus the search and order cost updates efficiently [11]. The decision of the next node expanded, the *successor*, is based on the *evaluation function*, $f(n)$: estimated cost of the cheapest solution through node $n$ (Note: lowercase $f$ is used here to specify the cost function at an individual search step, whereas capital $F$ was used previously to denote the cost function for an entire path). Informed search methods benefit from a *heuristic function* $h(n)$: the estimated cost of the cheapest path from a node $n$ to the goal state. Greedy best-first search is built solely on this heuristic, where $f(n) = h(n)$, expanding the node closest to the goal at each search step. The incorporation of the heuristic into the path cost makes the search algorithm more efficient because the search is focused in the direction of the robot, reducing the total number of state expansions [12].

The A* algorithm is perhaps the most popular best-first search method, adding to the heuristic the cost to reach the node, $g(n)$. That is, $f(n) = h(n) + g(n)$. The search algorithm, looking for the cheapest path, tries (expands) the node with the lowest $f(n)$ [13], [14]. To determine the optimal sequence of waypoints, the A* algorithm is a favorite for route search problems [15], [16], guaranteeing optimal solution paths [13]. Norvig and Russel [13] explain how A* is *optimally efficient*: no other optimal algorithm is guaranteed to expand fewer nodes than A*. As long as a better-informed heuristic is not used, A* will find the least-cost path solution at least as fast as any other method. The A*-PO algorithm defined by Lavin [10] builds off this optimality guarantee to offer an algorithm yielding Pareto optimal solution paths, specifically for problems of multiple path costs.

A* is a favorite method for path planning in static environments, where the positions of all obstacles and objects are fixed and known. The *dynamic* environment, on the other hand, may have obstacles and objects which vary positions with time. Similarly, an unknown or partially known environment calls for dynamic path planning because more is learned as the mobile robot progresses through the environment; the robot may receive info through e.g. an on board or offboard sensor. Thus it is important the optimal path(s) can be replanned when new information arrives.

The control architecture in mobile robotics is typically a combination of local and global planners, organized as shown in Fig 2. The reactive level handles local information, with real-time constraints. The deliberative, or global, level considers the entire world, likely requiring computation time proportional to the problem size [13]. The A* and D* algorithms, and their variations discussed in this paper, are global path planners.

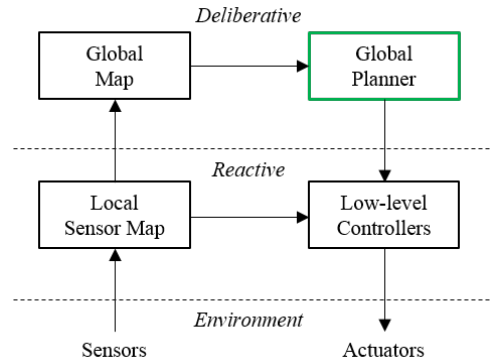

Fig. 2. High-level block diagram of the standard hybrid control system architecture for mobile robots [6]. The focus here is global path planning.

For reactive, or real-time planning, computational speed is a priority. Previous studies [17], [18] have modified A* for fast planning. The D* algorithm is a dynamic version of A*, built to be capable of fast rerouting when the robot encounters new obstacles in the environment [4]. D* has been shown to be up to two orders of magnitude more efficient than planning from scratch with A* [11]. The speed of these searching algorithms is increased dramatically, but at the cost of sub-optimal solution paths [6]. The algorithm defined here, D*-PO, yields Pareto optimal solutions with D* functionality.

*B. Pareto Optimality*

The MOO problem presents multiple cost criteria, where a solution stronger for one criterion may be weaker for another. There are two general approaches to optimizing for multiple objectives: (i) combine the individual objective functions into one composite function, and (ii) determine a *Pareto optimal* solution set. The first can be accomplished with weighted sums or utility functions, but selection of parameters is difficult because small perturbations in the weights can lead to very different solutions. These approaches also negate the optimality guarantees in select search algorithms, i.e. in A*. Despite these difficulties it is common the path cost function sums over the cost criteria at each step; the A* algorithm sums $h(n)$ and $g(n)$.

The second MOO approach finds the Pareto optimal set of the population, which is a set of solutions that are *non-dominated* with respect to each other. Non-dominated solutions are those in which there exist no other solutions superior in all attributes (i.e. objectives). Moving between Pareto solutions, there is always sacrifice in one objective to achieve gain in another objective [19]. In objective space, the set of non-dominated solutions lie on a surface known as the *Pareto front*. Fig. 3 illustrates the two-dimensional case, where there is a tradeoff between minimizing both functions $g_1$ and $g_2$. The second flowchart block of Fig. 1 shows the projection of path nodes into objective space.

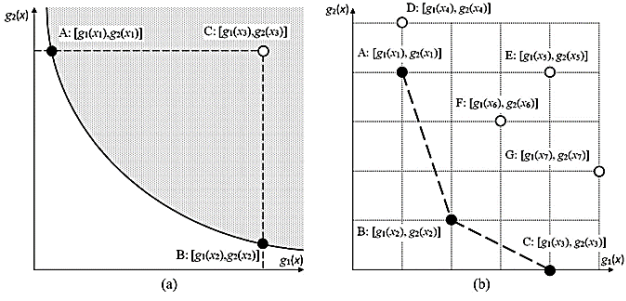

Fig. 3. Illustration of the Pareto front in two-dimensional objective space for (a) continuous and infinite, and (b) discrete and finite problems [20].

The MOO problem is defined as

$$\min_x [g_1(x), g_2(x), \ldots, g_N(x)] \quad (8)$$

subject to

$$h_I(x) \leq 0 \quad (9)$$
$$h_E(x) = 0 \quad (10)$$
$$x \in \Omega_X \quad (11)$$

where $g_i$ is the $i$ th objective function of $N$ total objectives; in path planning, $N$ is the number of cost criteria. The inequality and equality constraints are $h_I$ and $h_E$, respectively, and $x$ is the vector of optimization or decision variables in the set of $\Omega_X$.

Then a solution $x^*$ is Pareto optimal such that there exists no $x$ that makes

$$g_i(x) \leq g_i(x^*) \text{ for all } i = 1, \ldots, N \quad (12)$$
$$g_i(x) < g_i(x^*) \text{ for at least one } j \in [1, \ldots, N] \quad (13)$$

and the projection of $x^*$ in the objective space, i.e. the point $[g_1(x^*), g_2(x^*), \ldots, g_N(x^*)]$ is a Pareto point.

Multiobjective optimization is, therefore, concerned with the generation and selection of non-inferior solution points – those on the Pareto front. Pareto optimality is a crucial concept for finding solutions to MOO problems because identifying a single solution that simultaneously optimizes across several objectives is a nontrivial task [20].

MOO problems are typically handled with evolutionary algorithms, where it is common to use Pareto fronts in the fitness functions. The non-dominated paths are favored in the population, and this increases generation over generation [21]. The main drawback, however, of using evolutionary algorithms for path planning is computational complexity; the methods must generate populations (search space) composed of full or segmented paths. And the solution paths are suboptimal.

One may also consider that summing over the costs to calculate a composite $f$ presents another possible issue in search algorithms: depending on the current development of the path, some cost criteria may be favored over others, and this changes as the path development continues. For instance, the heuristic values – the estimated cost of the cheapest path from the current cell to the goal cell – will contribute more to the composite cost function close to the start than they will close to the goal. That is, near the start state $h(n)$ will have a greater influence on $f$ than will $g(n)$, and vise-versa for the goal state. Thus, as the path develops from start to goal, the heuristic value will contribute less and less. Using a Pareto front solves this issue because each cost criterion is valued as its own dimension in the Pareto space, not summed together.

III. TECHNICAL APPROACH

*A. Costmap*

To calculate cost functions at each step the search algorithms use a *costmap*. This representation of the configuration space is built off of the aforementioned occupancy grid, but now a cost value is assigned to each cell. Traversing a free space adds a unit cost to the path total, and the obstacles are represented by infinite cost; thus, they are not traversable. This is illustrated in the front image of Fig. 4. If traversing straight across a cell carries a unit distance cost, the cost for traversing a cell at a diagonal (a 45° angle) carries a cost of $\sqrt{2}$.

Yet this costmap only reflects the distance of taking a given path through the configuration space. For a MOO problem, the path cost needs to consider the other cost criteria, for which we use additional *layers*. Each additional cost layer adds a dimension to the Pareto space, from which the Pareto front is calculated.

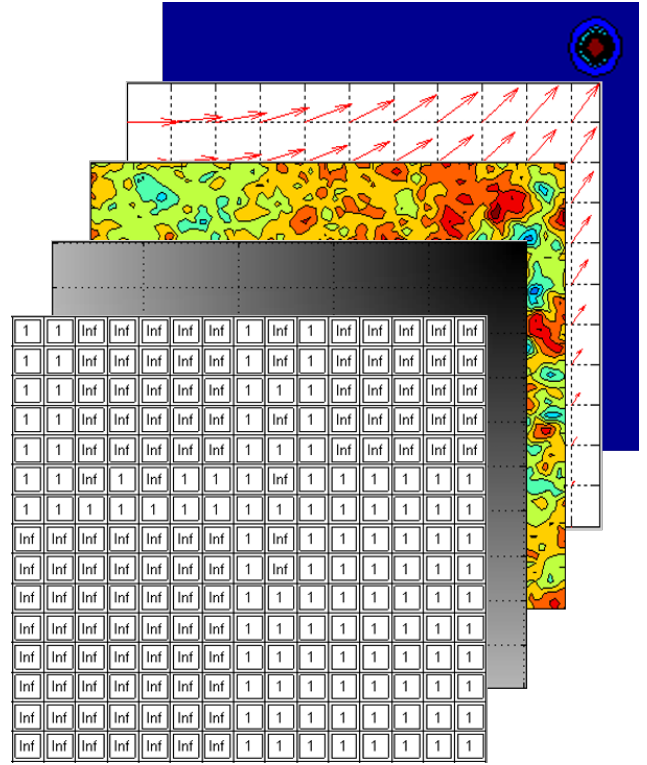

Fig. 4. Illustration of path cost layers for the criteria of this study. The individual layers represent (left to right) the occupancy grid, the heuristic, the elevation, the solar vectors, and the risk.

The first costmap layer is the distance cost, $g(n)$, and the second layer is the heuristic, $h(n)$. These two suffice for traditional A* and D* search, but we're interested in optimizing the robot's path for additional criteria. A third layer, $e(n)$, is added to the costmap to incorporate the elevation costs. The fourth layer, solar deviation $s(n)$, is dynamic, changing at each step. That is, to simulate solar rays changing direction over time, the solar vector is rotated as the robot progresses. The fifth layer, $r(n)$, has a risk value associated with each location in the workspace. Fig. 4 illustrates the layering of these path costs. An $n$-dimensional Pareto space has at most $n$ layers to the costmap. The points on the Pareto front are non-dominated across the $n$ dimensions, one dimension for each cost.

The cost values associated with the layers makeup a given node's projection to objective space, as in the second flowchart block of Fig. 1. All nodes in the open list are projected to objective space. At each search step the costs for each objective are normalized [0:1] over the nodes in the open list. This is an unnecessary step, however, for D*-PO (and A*-PO) because each cost value is relative to the cost metric's dimension in Pareto space. But normalization is necessary for the A* and D* algorithms because the composite cost function merges costs of varied measuring units and scales; without normalization, the objective values would not carry equal weight in the combined objective function. For instance, the elevation values are small relative to the distance values, and without normalization the elevation metric would be insignificant

### B. D*-PO Search Algorithm

The algorithm presented in this study, D*-PO, is effectively the standard D* search algorithm but for a key modification: rather than computing the cost function $f$ by summing cost criteria, D*-PO calculates the Pareto front of the cost criteria. The original D* paper by Stentz [12] explains D* extensively, including the algorithm's pseudocode, and is not repeated here. The inherent dynamic qualities of D* are maintained in D*-PO. The only departure from D* is using the Pareto front of the priority queue to determine the next successor node, as shown in the second through fourth flowchart blocks of Fig. 1. The D* search algorithm is guaranteed complete, but not optimal. The D*-PO algorithm, however, is guaranteed to yield complete and Pareto optimal solution paths.

Calculating the Pareto front of the open list will at times yield multiple Pareto points. To break the tie amongst the Pareto optimal successor nodes, a *priority cost* is defined. This cost is customizable, and can be e.g. a preferred criterion or a new criterion. As shown in the final flowchart block of Fig. 1, the cheaper node according to this priority cost is chosen for expansion. Because the priority cost decides between Pareto optimal nodes, the D*-PO search algorithm still maintains the quality that every step in the solution path is Pareto optimal.

## IV. RESULTS

The MOO path planning algorithms were tested in simulated mobile robot environments. The computer simulation environment included a Lenovo notebook computer with Intel Core i5 vPro CPU and 4 GB memory, running on Windows 8.1. The code is written in MATLAB R2013a.

### A. Algorithm Comparison

The algorithms yield the same solutions for single path cost problems, as expected. Where the advantages of D*-PO are significant is for MOO path problems. D*-PO was evaluated by comparing it with A* and D* for a set of 100 simulated environments.

The workspaces were setup as a 100x100 cell grids of randomly assigned obstacles, sized large, medium, and small. On average, the obstacles accounted for 23% of the configuration space. The start and goal locations were fixed at the lower-left (0,0) and upper-right (100,100), respectively. For each workspace a random 100x100meter terrain was sampled from HiRISE Mars terrain data [22], where the elevations were normalized [0:1] for each sample. Fig. 5 shows ten sampled workspaces from one Mars image. The solar vector is defined to rotate counterclockwise one radian for every path step, initially set at [1,0]. For each workspace a random cell was chosen as the location of max risk, with the risk decreasing radially with the inverse of the distance squared, as described in equation (7).

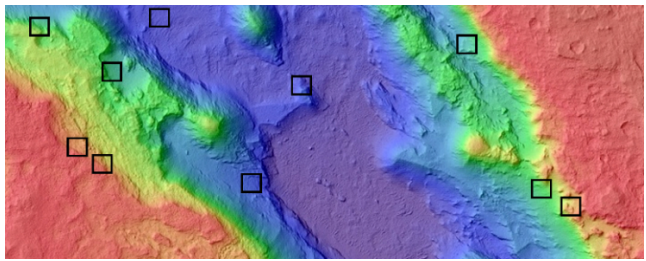

Fig. 5. Layered bedrock northwest of the Hellas Region of Mars [23]. The black squares are 100x100m regions randomly selected for simulation workspaces.

The optimization objectives, as presented above in (4)-(7), were to minimize the total path distance, elevation, solar vector deviation, and proximity to risk. Fig. 6 shows two examples of simulation runs with the resulting paths for the A*, D*, and D*-PO search methods, where the red and green marked squares represent the start and goal states, respectively. The left side diagrams of Fig. 6 are the final solution paths plotted over the layer of obstacles (red) and free spaces, where the background layer represents the heuristic cost. The right side diagrams show the same paths over a contour map, representing the elevation layer of the costmap. These plots also include the solar vector at the beginning and end of the paths.

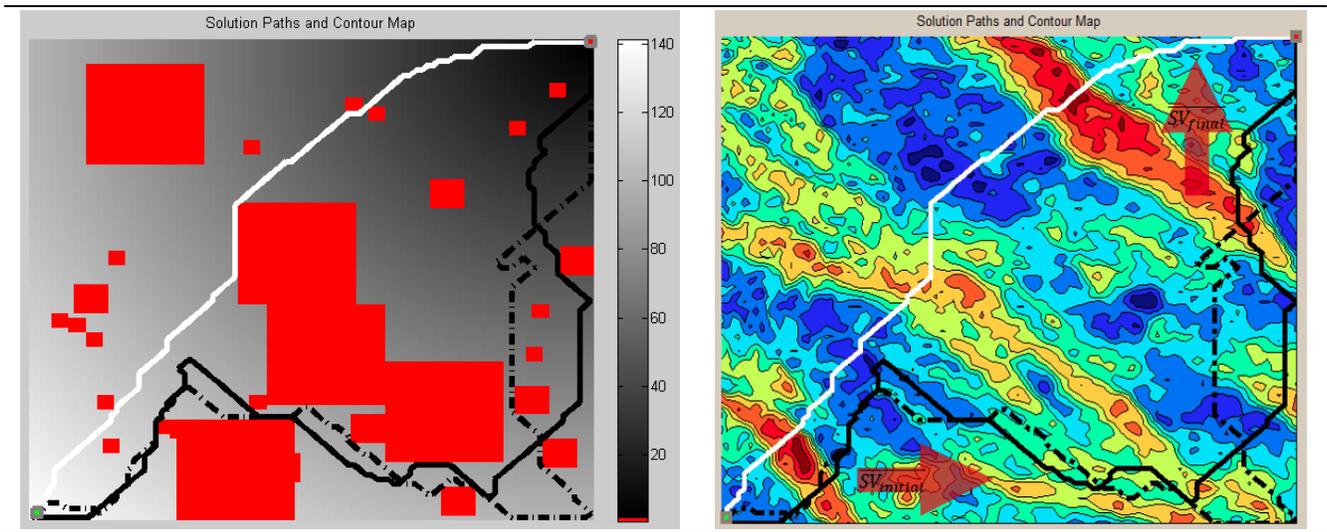

| Algorithm | Length (m) | Elevation [0:1] | Solar Deviation | Risk Proximity (m) | Computation Time (sec) |
|---|---|---|---|---|---|
| **A*** | 254.18 | 0.383 | 0.107 | 0.00239 | 6.47 |
| **D*** | 231.59 | 0.418 | 0.333 | 0.00159 | 24.64 |
| **D*-PO** | 155.58 | 0.364 | 0.108 | 0.00029 | 31.19 |

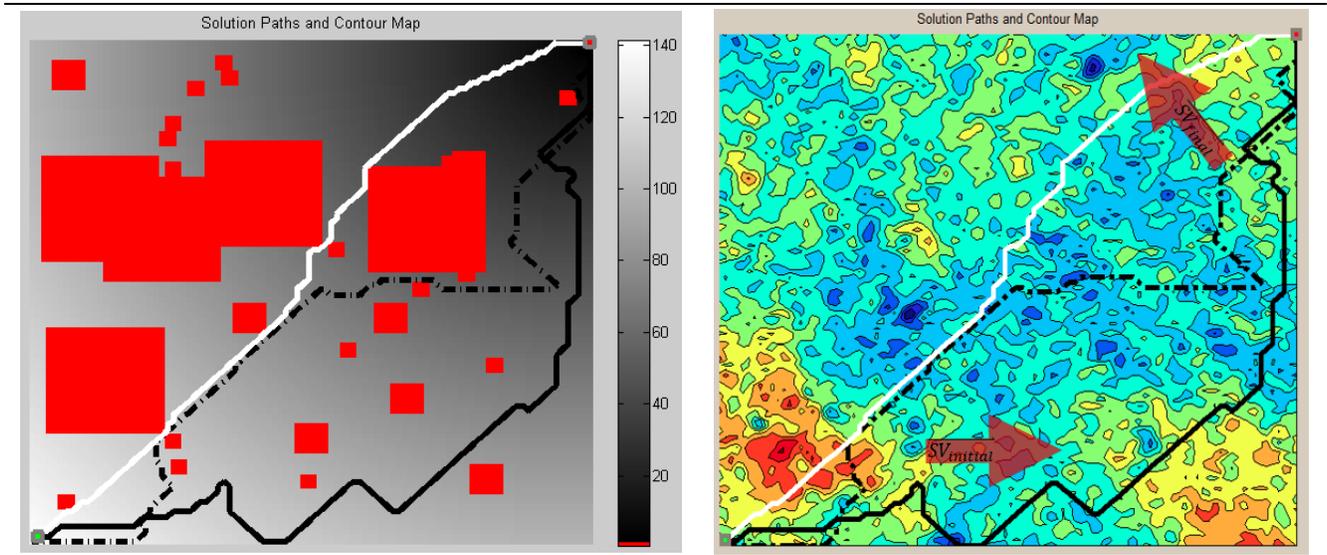

| Algorithm | Length (m) | Elevation [0:1] | Solar Deviation | Risk Proximity (m) | Computation Time (sec) |
|---|---|---|---|---|---|
| **A*** | 188.48 | 0.522 | 0.255 | 0.00059 | 7.25 |
| **D*** | 201.97 | 0.54 | 0.229 | 0.00252 | 12.74 |
| **D*-PO** | 149.14 | 0.488 | 0.179 | 0.00051 | 14.12 |

Fig. 6. Two examples of the 100 simulation runs; plots in the same rows show the same solution paths over different cost layers. The A*, D* and D*-PO paths are plotted with black dashed, black, and white lines, respectively. The path background of the left figures represents the heuristic and occupancy grid, with the colorbar showing distance to the goal in meters. The right figures show the contour map used for the elevation cost, and the solar vectors at the start and goal states. The solution path metrics for these individual simulations are shown in the tables below the figures.

TABLE I. SIMULATION RESULTS

| | Algorithm Comparison Metrics (mean values) | | | | Computation Time (sec) |
|---|---|---|---|---|---|
| | $F_1$: Length (m) | $F_2$: Elevation [0:1] | $F_3$: Solar Deviation | $F_4$: Risk Proximity (m) | |
| **A\*** | 213.69 | 0.476 | 0.0924 | 0.263 | 5.17 |
| **D\*** | 207.09 | 0.483 | 0.4181 | 0.439 | 26.36 |
| **D\*-PO** | 154.75 | 0.474 | 0.0758 | 0.161 | 32.36 |

For the sample workspaces in this example, it is clear to see the benefits of calculating the Pareto front at each search step. The data over the set of 100 simulations echo these results, as shown in Table 1. Lower values are preferred for all metrics. The D*-PO algorithm outperforms both A* and D* across all of the MOO criteria, even though the resulting solution paths are not completely non-dominated solutions – i.e. at some steps the solution is the preference of the non-dominated frontier. Thus, D*-PO solutions are always at least as good as those produced by A* and D*. The search time results show paths with Pareto optimal steps can be obtained efficiently with the D*-PO algorithm; A* is markedly faster than the others because the dynamic features of D* rely on priority queues that reopen (raise) states. All algorithms gave complete solution paths for all 100 simulations.

The average elevation of each solution path is used as a metric to compare the robot's net incline from start to goal. A path of a given average elevation implies the robot traversed up less slope (or down more slope) as compared to a path of higher average elevation. All paths share common start and goal states, so the elevation values are relative to one another.

The four functions cover the five cost criteria because both the distance travelled and the heuristic contribute to $F_1$. The simulations show the D*-PO algorithm provides the least-cost global path according to several independent preferences for a mobile robot in practice – e.g. planetary exploration rover.

Expanding on these Mars rover simulations, one may account for more elaborate thermal constraints, such as heating of sensitive components by direct sunlight. Minimizing the number of turns could be another optimization objective.

## V. CONCLUSION

In this study, global path planning for mobile robots is investigated. The optimal path is generated according to several cost criteria, solving the multiobjective optimization problem with the presented D*-PO algorithm. As demonstrated in the previous section, D*-PO is capable of providing paths where each step is Pareto optimal, and computes these solutions efficiently. In comparison to the traditional D* algorithm, it can be concluded the incorporation of Pareto fronts in D*-PO offers a better MOO search algorithm – a method that is both dynamic and optimal. D*-PO also is shown to outperform the widely-used A* search for MOO problems.

This study used a cell grid representation, which reduces computational complexity at the expense of reducing completeness. That is, solution paths are restricted to a constrained mobility. Future implementations of D*-PO (and A*-PO) may opt for *state lattice* representation: a discretized set of all reachable configurations of a system [23]. This representation would better model mobility constraints, leading to superior motion planning results because no time is wasted either generating, evaluating, or fixing infeasible plans.

Continuation of this study may aim to further validate the D*-PO and A*-PO algorithms by evaluating paths based on a complete mobile robot energy expenditure metric. For a given path, this metric will provide a single energy value which would include the robot's energy output (factors of distance traversed, elevation change, and degrees of turns) and input (from solar array). Additionally the next implementation may use Pareto fronts in D* Lite which has been found to be slightly more efficient for some navigation tasks and easier to analyze [24].

The D*-PO and A*-PO algorithms would be beneficial to many applications for mobile robots with global path planning, including agricultural harvesting and information gathering (i.e. drones), disaster relief, DARPA challenges, factory and residential robot workers, and exploration rovers. D* is a very general algorithm and can be applied to problems in artificial intelligence other than robot motion planning, essentially any path cost optimization problem where the cost parameters change during the traverse of the solution [12]. D*-PO should be up to the task as well.